\documentclass[runningheads]{llncs}

\usepackage[T1]{fontenc}

\usepackage{cite}
\usepackage{booktabs}
\usepackage{multirow}
\usepackage{graphicx}
\usepackage{subfigure}
\usepackage{amssymb}
\usepackage{url}
\usepackage[linesnumbered,ruled,vlined,noend]{algorithm2e}
\usepackage{amsmath}
\usepackage{pdflscape}

\title{Beyond Predefined Actions: Integrating Behavior Trees and Dynamic Movement Primitives for Robot Learning from Demonstration}
\titlerunning{Integrating BTs and DMPs for Robot Learning from Demonstration}
\author{David C\'aceres Dom\'inguez \orcidID{0000-0003-2279-9418}
\and Erik Schaffernicht \orcidID{0000-0002-0804-8637}
\and Todor Stoyanov \orcidID{0000-0002-6013-4874}
}%
\authorrunning{D. C\'aceres Dom\'inguez et al.}
\institute{Center for Applied Autonomous Sensor Systems (AASS), Örebro University, Sweden
\email{\{firstname.lastname\}@oru.se}}

\begin{document}
\maketitle

\begin{abstract}

Interpretable policy representations like Behavior Trees (BTs) and Dynamic Motion Primitives (DMPs) enable robot skill transfer from human demonstrations, but each faces limitations: BTs require expert-crafted low-level actions, while DMPs lack high-level task logic. 
We address these limitations by integrating DMP controllers into a BT framework, jointly learning the BT structure and DMP actions from single demonstrations, thereby removing the need for predefined actions. 
Additionally, by combining BT decision logic with DMP motion generation, our method enhances policy interpretability, modularity, and adaptability for autonomous systems. 
Our approach readily affords both learning to replicate low-level motions and combining partial demonstrations into a coherent and easy-to-modify overall policy. 
\keywords{Behavior Trees  \and Learning from Demonstration.}
\end{abstract}
\section{Introduction}
In robotics, there has been a long-standing pursuit to enable machines to learn tasks through human demonstrations~\cite{schaal1996learning, atkeson1997robot}. 
A key challenge remains in designing a flexible and interpretable framework that allows intelligent autonomous systems to execute complex behaviors efficiently.
Among promising approaches explored in recent years is the use of Behavior Trees (BTs)~\cite{DBLP:journals/corr/abs-1709-00084} --- hierarchical structures that offer an intuitive framework for representing and orchestrating a robot's decision-making processes.

Prior work has demonstrated the capacity to learn the structure of BTs from demonstrations~\cite{french2019learning}, however, existing methods require access to task-specific expert knowledge in the form of a predefined set of actions.
This requires manual effort and not only consumes considerable time, but also limits the range of behaviors that can be composed. 
In parallel, efforts to address the issue of designing low-level action primitives for BTs have explored using Dynamic Movement Primitives (DMPs)~\cite{liu2022robotic} and Motion Generators~\cite{rovida2018motion}. 
However, these methods require manually designing the BT structure and the manual integration of predefined actions, often necessitating action parameter adjustment during execution.
Thus, current approaches to learning BTs from demonstration are incomplete and limited to learning the BT structure when the actions are manually defined, or conversely, learning the action parameters when the BT structure is fixed.

\begin{figure}[t!]
\centering
\includegraphics[width=0.6\columnwidth]{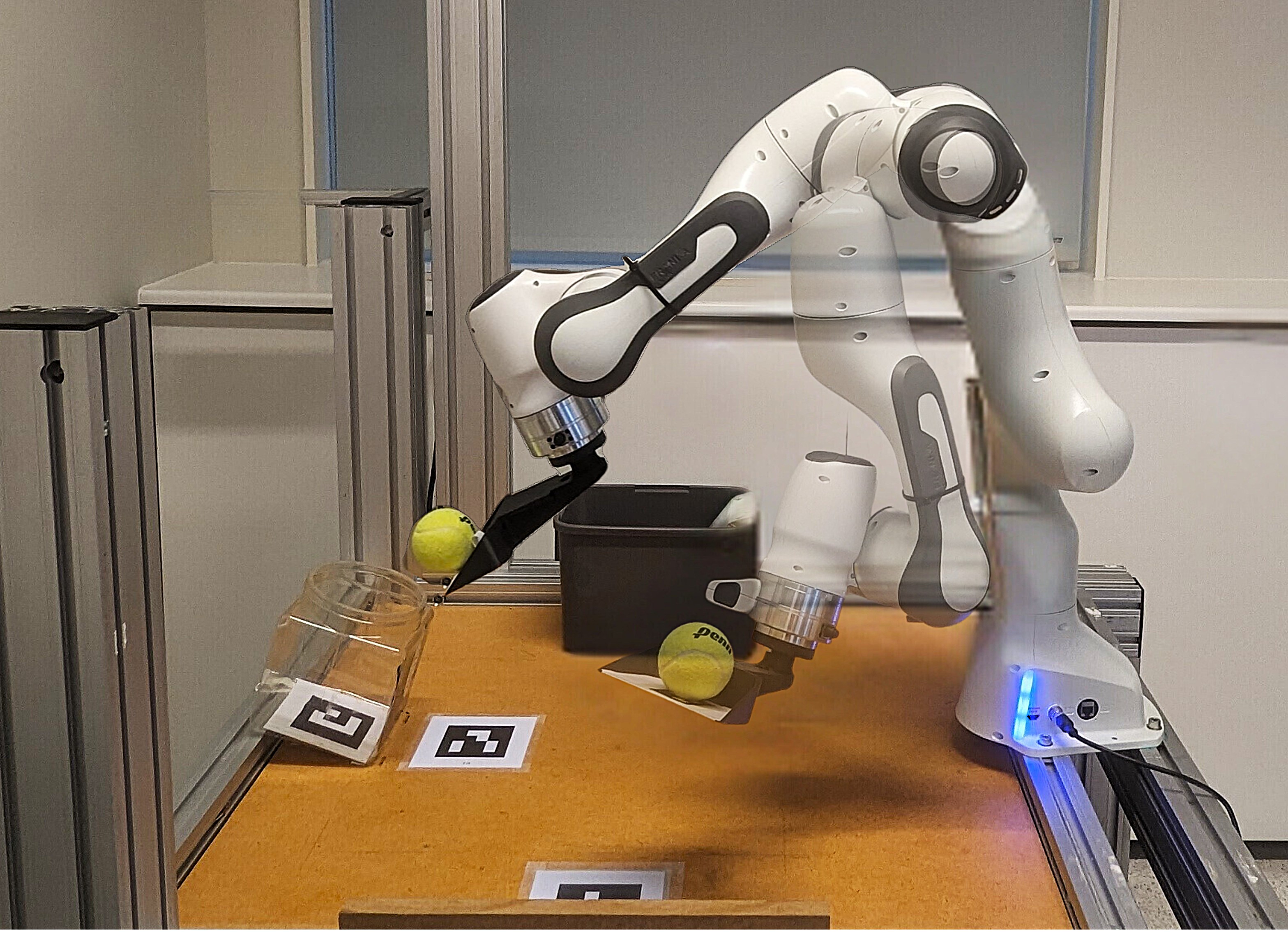}
\caption{Experimental setup for evaluation: a Franka Emika Panda robotic manipulator equipped with a shovel, scoops a tennis ball and deposits it into one of two stations.}
\label{fig:setup:demo}
\end{figure}

In this paper, we introduce a new approach for learning both the high-level task structure and associated low-level actions from demonstrations, employing BTs as a compositional structure and DMPs for low-level motion generation. 
We encapsulate each DMP within a BT action node, leveraging the pre- and post-conditions of the BT nodes to supervise execution. 
This allows us to create complex and adaptive behaviors that can handle different situations and contingencies. 
For example, we can use BTs to implement conditional branching, looping, and fallback mechanisms, which are not possible with DMPs, nor other imitation learning approaches. 

Our main contribution is a novel method that learns a comprehensive BT policy from human demonstration. 
Our approach seamlessly segments the demonstration movements into distinct BT actions and their associated condition nodes, enabling the automatic learning of the BT structure.
By leveraging DMPs, our method eliminates the need for predefined action nodes when learning the structure of a BT, allowing us to encapsulate a wide range of motions directly within the BT.
Additionally, the higher-level decision logic of BTs facilitates seamless switching between different DMPs and enhances the interpretability and modularity of the learned robot policy.

\section{Background}

\subsection{Behavior Trees} \label{background_bts}
Behavior Trees are task-switching structures offering modularity, reusability, and reactivity, serving as an alternative to Finite State Machines~\cite{iovino2024comparison}. 
A BT consists of a root node, control nodes, and leaf nodes~\cite{DBLP:journals/corr/abs-1709-00084}. 
Execution begins at the root, sending "tick" signals through the tree to activate nodes. 
Nodes execute only when ticked, returning one of three statuses: \textit{Running}, \textit{Success}, or \textit{Failure}.
\begin{figure}[t!]
    \centering
    \subfigure[]{
        \includegraphics[trim={0 0.3cm 0 0.3cm},clip,height=0.06\linewidth]{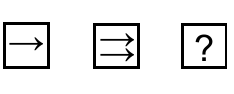}
        \label{fig:control-nodes}}
    \subfigure[]{
        \includegraphics[height=0.055\linewidth]{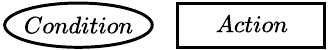}
        \label{fig:execution-nodes}
        }
  
    \caption{Behavior Tree nodes:~\subref{fig:control-nodes} Control nodes (Sequence, Parallel, Fallback) and~\subref{fig:execution-nodes} Execution nodes (Condition, Action).}
\end{figure}

\noindent Control Nodes (Fig. \ref{fig:control-nodes}) include Sequence, Fallback, and Parallel types. 
A Sequence ticks children left-to-right, succeeding only if all children succeed. 
A Fallback ticks children left-to-right until one succeeds. 
Both stop ticking further children if one returns \textit{Running}. 
A Parallel ticks all children simultaneously, succeeding if a set number succeed.
Leaf Nodes (Fig. \ref{fig:execution-nodes}) are either Action or Condition nodes. 
An Action node executes commands when ticked, returning \textit{Running}, \textit{Success}, or \textit{Failure} based on progress or outcome. 
Condition nodes evaluate propositions, returning \textit{Success} if true and \textit{Failure} otherwise

\subsection{Dynamic Movement Primitives} \label{DMP_section}
Dynamic Movement Primitives (DMPs) are a flexible framework for learning trajectories from demonstration, representing both periodic and discrete movements. 
In this study, we build on the DMP framework introduced in~\cite{GINESI2021103844}, based on the formulation in~\cite{5152423, 4755937}. 
These primitives comprise a system of second-order ordinary differential equations, 
resembling a mass-spring-damper system with an added forcing term. 
DMPs model the forcing term to generalize trajectories to new start/goal positions while preserving their learned shape.
The formulation for one-dimensional DMPs is outlined as:
\begin{equation} \label{eq:one_dmp}
\begin{cases}
\begin{aligned}
    \tau \dot{v} &= K(g - x) - Dv + K(g - x_0) s +Kf(s) \\
\end{aligned} \\
\begin{aligned}
    \tau \dot{x} &= v ,
\end{aligned}
\end{cases}
\end{equation}

\noindent where \( x \), \( v \), \( x_0 \), \( g \in \mathbb{R}\) represent the position, the velocity, the initial position, and the goal, respectively. 
The constants \( K \), \( D \) are the spring and damping terms, chosen in such a way that the associated homogeneous system is critically damped: \( D = 2 \sqrt{K}\). 
\( \tau \in \mathbb{R^+}\)  is a temporal scaling factor, and \( f \) is a real-valued nonlinear forcing (also called perturbation) term, defined as:
\begin{equation}
f(s) = \frac{\sum_{i=1}^{N} \omega_i \psi_i(s)}{\sum_{i=1}^{N}\psi_i}, 
\end{equation}
with Gaussian functions $\psi_i(s) = \exp{(-h_i(s-c_i)^2)}$ with centers $c_i$ and widths $h_i$. 
Instead of time, $f$ depends explicitly on a phase variable $s$, 
\begin{equation}\label{eq:canonical_system}
    \tau \dot{s} = -\alpha s,
\end{equation}
where $\alpha$ is the predefined constant for the canonical system defined in equation~\ref{eq:canonical_system}.
The learning process consists of determining the weights $\omega_i \in \mathbb{R}$ by computing $f(s)$ for a given desired trajectory.
DMPs support chaining, switching, and blending, enabling transitions without exact starting states and allowing demonstrations to be segmented into independently learned DMPs.

\section{Related work} \label{related_work}
\subsection{Learning from Demonstration}
Interpretable policies are crucial for ensuring that human operators can understand, analyze, and debug learned robotic behaviors.
Traditional Learning from Demonstration (LfD) methods, such as those based on Hidden Markov Models~\cite{calinon2010learning}, neural networks~\cite{duan2017one}, or DMPs ~\cite{matsubara_learning_2011,kulvicius_joining_2012} have shown to be proficient at encoding and reproducing complex behaviors. 
However, they are black-box methods that learn complex mappings from input to output without an explicit representation in a human-understandable form.
They suit tasks where accurate output matters more than understanding the decision-making process.

Caccavale et al.~\cite{caccavale_kinesthetic_2019} propose a method combining Hierarchical Task Networks with Dynamic Motion Primitives (DMPs) to structure complex tasks into smaller sub-tasks while using DMPs for motion generation. 
However, their approach requires manually defining abstract task descriptions, which is both time-intensive and limits adaptability to unstructured tasks.

Other works synthesize human-readable programs mapping directly to robot actions, enhancing transparency~\cite{patton2024programming, trivedi2021learning}.
However, they rely on manually defined task or action abstractions, which can be a bottleneck, especially in complex tasks where these abstractions may not cover all necessary behaviors. 
To the best of our knowledge, there are currently no baseline interpretable LfD methods that can both abstract from low-level actions and extract high-level decision logic.

\subsection{Learning BTs from demonstration}
Learning BTs has been a subject of extensive research, with a wide range of techniques aimed at automatically generating, enhancing, or optimizing their structure.
A variety of approaches have been proposed, including Genetic Programming~\cite{8319483}, Reinforcement Learning~\cite{mayr2021learning} and Case-Based Reasoning~\cite{palma2011extending}.

In the context of learning BTs from demonstration, methods typically rely on intermediary algorithms like CART~\cite{breiman2017classification} or C5.0~\cite{quinlan1986induction} to initially construct a Decision Tree (DT) from a labeled dataset and then apply a conversion algorithm to derive a BT. 
These techniques leverage the similarities between logical expressions derived from the resulting DT and specific BT nodes.
However, describing reactive behaviors using DTs often requires repeated predicate re-evaluation at varying tree depths, resulting in larger and less interpretable BTs~\cite{wathieu2022re}. 
To address this challenge, efforts have been made to optimize the DT's logical statements. 
Sagredo-Olivenza et al.~\cite{sagredo2017trained} rely on manual optimizations, while French et al.~\cite{french2019learning} and Gugliermo et al.~\cite{10105979} propose logic minimization algorithms to reduce the DT's logical statements prior to conversion. 
Building on the work of French et al.~\cite{french2019learning}, Wathieu et al. \cite{wathieu2022re} introduce the RE:BT-Espresso algorithm, which enhances the interpretability of the learned BT by eliminating logical redundancies. 

All current BT learning methods rely on the availability of a labeled dataset with pre-defined actions, thereby limiting BT actions to either a manually engineered set of complex action nodes or to using discretized low-level atomic actions. 
Both of these options have undesirable consequences: the former are limiting in the expressivity and variety of motions and time-consuming to define; the latter results in large and hard-to-interpret BTs. 
In contrast, our approach uses DMPs to segment, label, and learn action nodes from human demonstrations, assembling them into interpretable BTs that overcome predefined action set limitations and enhance flexibility when learning the structure.
\section{Proposed method} \label{proposed_method}
\begin{figure}[t]
    \centering
    \includegraphics[width=1\linewidth]{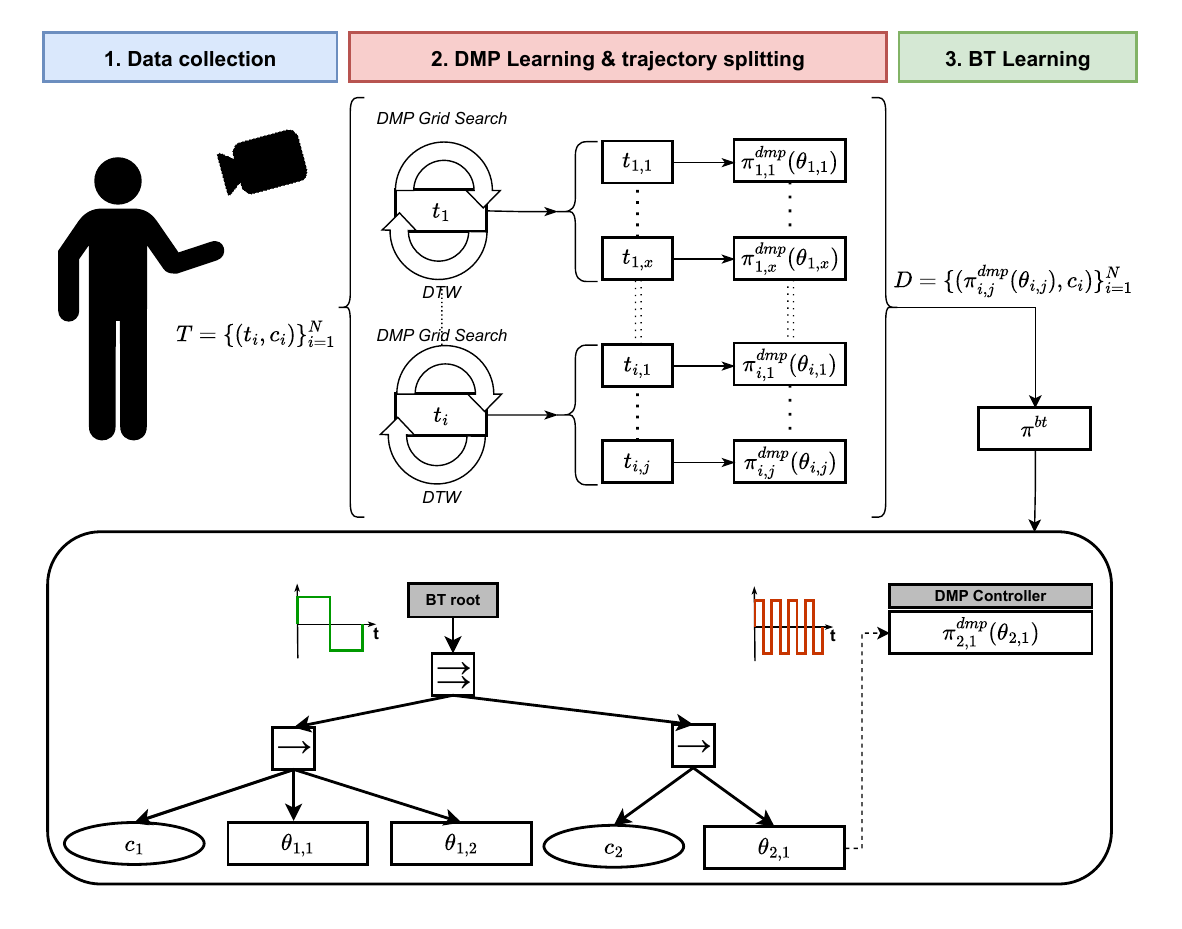}
    \caption{Proposed method flowchart. 
    (1) Data collection gathers demonstrations $T$, linking trajectories $t_i$ with state variables $c_i$. 
    (2) DMP Learning recursively fits segmented trajectories with DMPs $\pi^{dmp}_{i,j}(\theta_{i,j})$, evaluated via DTW. 
    (3) BT Learning trains a Decision Tree with CART and converts it to a BT policy $\pi^{bt}$ using RE:BT-Espresso.}
    \label{fig:flowchart}
\end{figure}

The main contribution of our approach is the integration of BTs and DMPs to learn robot policies from human demonstrations without requiring manual labeling or predefined BT actions.
As illustrated in Fig.~\ref{fig:flowchart}, our methodology for learning BTs with DMPs involves three main stages. 
Sec.~\ref{combining_bts_dmps} introduces the framework, while Sec.~\ref{data_collection}-\ref{bt_learning} provide detailed explanations of each stage.

\subsection{Combining DMPs with BTs} \label{combining_bts_dmps}
Given a dataset $T=\{(t_i, c_i)\}^{N}_{i=1}$ containing demonstrated trajectories $t_{i}$ and associated state variables $c_i$, the objective is to learn an upper-level BT policy $\pi^{bt}$ and a set of lower-level DMP control policies $\pi^{dmp}_{i,j}(\theta_{i,j})$ such that:

\begin{equation}
    \pi^{bt} : (c_i, s_{bt}) \to \pi^{dmp}_{i,j}(\theta_{i,j})
\end{equation}

\noindent where $s_{bt}$ represents the current state of execution of the BT policy $\pi_{bt}$ determined by the current location of the tick. 
$\theta_{i,j}$ represents the parameters of the DMP policy for segment $j$ of trajectory $i$. 
Each DMP policy is encapsulated within a BT Action node.
Both policies --- $\pi^{bt}$ and $\pi^{dmp}_{i,j}$ --- operate concurrently but at different frequencies, with $\pi^{dmp}_{i,j}(\theta_{i,j})$ commanding the robot's low-level trajectory, denoted as $\tau_{i,j}$.
Our approach learns the set of $\pi^{dmp}_{i,j}(\theta_{i,j})$ from $T$ using the trajectory splitting strategy detailed in Fig.~\ref{fig:recursive-segmentation}.
The high-level BT policy $\pi^{bt}$ is learned from the dataset $D=\{(\pi^{dmp}_{i,j}, c_i)\}^N_{i=1}$ using RE:BT-Espresso~\cite{wathieu2022re}.
\subsection{Data collection} \label{data_collection}
The first stage collects a dataset of demonstrations $T=\{(t_i, c_i)\}^{N}_{i=1}$, where each demonstration comprises the observed trajectories $t_i$ and associated state variables $c_i$. 
These demonstrations can be obtained from a variety of sources, including direct (e.g. Kinesthetic teaching) and indirect (e.g., visual systems) teaching. 
The state $c_i$ may include environmental variables (e.g., object pose, sensor readings) or engineered conditions (e.g., gripper status).
If an engineered condition changes during a demonstration, a new trajectory $(t_{i+1},c_{i+1})$ is recorded and added to $T$. 
Associating $c_i$ to each $t_i$ ensures that each learned $\pi^{dmp}$ coincides with the subset of $c_i$ that will later be used in $\pi^{bt}$, resulting in continuity in the trajectories between states. 
Additionally, it allows the identification of different $t_i$ from similar demonstrations that belong to the same $\pi^{dmp}$.

\subsection{DMP Learning and trajectory-splitting} \label{dmp_learning}

Learning DMPs involves several parameters $\theta$ --- e.g., the number of basis functions, the width of the basis functions, and the constant of the canonical system. Consequently, we perform a grid search over $\theta_i$ in $\pi^{dmp}_i(\theta_i)$, ultimately selecting the optimal parameter configuration that best fits $t_i$.
\begin{figure}
    \centering
    \includegraphics[width=0.9\linewidth]{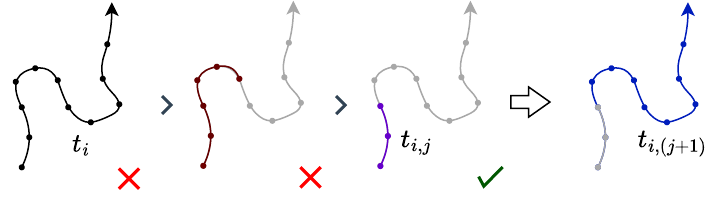}
    \caption{Recursive segmentation of $t_i$: split into $t_{i,j}$ if $DTW(t_{i,j}, \tau_{i,j}) > \epsilon$. Valid segments (green) link to DMPs when $DTW \leq \epsilon$, continuing on $t_{i,j+1}$.}  
    \label{fig:recursive-segmentation}
\end{figure}

To determine which $\pi^{dmp}_i(\theta_i)$ best fits to the demonstration trajectory 
$t_i=\{t_{i_1},t_{i_2},...,t_{i_n},...,t_{i_{|t_i|}}\}$, 
we measure the similarity between the demonstration and the generated trajectory 
$\tau_i=\{\tau_{i_1},\tau_{i_2},...,\tau_{i_m},...,\tau_{i_{|\tau_i|}}\}$. 
We approximate trajectory similarity using FastDTW~\cite{salvador2007toward} --- an approximation of Dynamic Time Warping (DTW)~\cite{kruskall1983symmetric} that has a linear time and space complexity. 
The basic idea is to find the distance between the two sequences, which may vary in length or speed, by warping their time axes and constructing the warp path $W$:
\begin{equation}
    W=\{w_1,w_2,...,w_k\}   \qquad     max(|t_i|,|\tau_i|)\leq K < |t_i|+|\tau_i|
\end{equation}

\noindent where $K$ is the length of the warp path and the $k$-th element of the warp path is $w_k=(n,m)$.
We calculate the DTW distance between $t_i$ and $\tau_i$ as:
\begin{equation}
    DTW(t_i, \tau_i) = \sum^K_{k=1}c(t_{i_n}, \tau_{i_m})
\end{equation}
\noindent where $c(t_{i_n}, \tau_{i_m})$ is the Euclidean distance between the two data point indexes (one from $t_i$ and one from $\tau_i$) in the $k$-th element of the warp path.
We measure DTW distance for each of the generated $\pi^{dmp}_i(\theta_i)$ during the grid search and select the one with the lowest value.
Afterwards, we perform the following evaluation based on a fitness threshold $\epsilon$: 

\begin{itemize}
    \item If $DTW(t_i, \tau_i) \leq \epsilon$, $\pi^{dmp}_i(\theta_i)$ is deemed successfully learned, and the trajectory $t_i$ is labeled in the dataset with that particular DMP.
    \item If $DTW(t_i, \tau_i) > \epsilon$, $\pi^{dmp}_i(\theta_i)$ is deemed unsuccessfully learned and $t_i$ is split in half following the strategy in Fig.~\ref{fig:recursive-segmentation}. 
    Then, we perform a grid search for each segment $t_{i,j}$ to find the best $\pi^{dmp}_{i,j}(\theta_{i,j})$. 
    This segmentation is recursively performed until $DTW(t_{i,j}, \tau_{i,j}) \leq \epsilon$ and all $t_{i,j}$ are labeled with a DMP.
\end{itemize}

\noindent The process produces the dataset $D = {(\pi^{dmp}_{i,j}, c_i)}^N_{i=1}$, detailed in Algorithm~\ref{algo:dmp_learning}.

\begin{algorithm}
\SetAlgoNoLine
\SetKwInOut{Input}{Input}
\SetKwInOut{Output}{Output}
\SetKwFunction{performGridSearch}{DMP\_grid\_search}
\SetKwFunction{evaluate}{DTW}
\SetKwFunction{segment}{segment}
\SetKwFunction{sample}{gen}
\SetKwComment{Comment}{\#}{}

\Input{Trajectories $T$, threshold $\epsilon$, minimum trajectory support $min\_split$}
\Output{Set of learned DMPs $D$}

\For{$t_i$ \textbf{in} $T$}{
    Initialize first segment index: $j \gets 0$\;
    Set active segment to full trajectory: $t_{i,j} \gets t_i$\;
    \While{$t_i > 0$ \text{and} $t_{i,j} > \text{min\_split}$}{
        Fit DMP to $t_{i,j}$: $\pi^{dmp}_{i,j}(\theta_{i,j})  \gets $\performGridSearch($t_{i,j}$)\;
        Sample DMP trajectory: $\tau \gets \sample(\pi^{dmp}_{i,j})$\;
        \If{$\evaluate(t_{i,j}, \tau) \leq \epsilon$ \textbf{or} $t_{i,j} \leq min\_split$}{
            $D \gets D \cup \{\pi^{dmp}_{i,j}(\theta_{i,j}), c_{i}\}$\;
            $t_i, t_{i,j} \gets (t_i \backslash t_{i,j})$\;
            $j \gets (j + 1)$\;
        }
        \Else{
            Split $t_{i,j}$ in half: $t_{i,j} \gets$ \segment($t_{i,j}$)\;
        }
    }
}
\caption{DMP learning and trajectory splitting}
\label{algo:dmp_learning}
\end{algorithm}

\subsection{Behavior Tree Learning}\label{bt_learning}
Once $D$ is learned, we take the associated environmental variables $c_i$ as conditions and associated $\pi^{dmp}_{i,j}(\theta_{i,j})$ as labels and feed them into the RE:BT-Espresso algorithm described in Sec.~\ref{related_work} to obtain $\pi^{bt}$.
The process first learns a decision tree (DT) by classifying $\pi^{dmp}_{i,j}(\theta_{i,j})$ based on a set of rules derived from $c_i$ and then converts it to a BT by using the RE:BT-Espresso algorithm~\cite{wathieu2022re}.
In particular, RE:BT-Espresso employs the CART~\cite{lewis2000introduction} algorithm to learn the DT from $D$ and subsequently transforms the DT into boolean equations, one for each $\pi^{dmp}_{i,j}(\theta_{i,j})$.
Then, RE:BT-Espreeso simplifies the boolean equations using logic minimization and constructs the BT policy $\pi^{bt}$ after pruning unnecessary nodes.

This methodology provides a robust and flexible approach to learning BTs from demonstration, and its combination with DMPs allows for the learning of complex behaviors from a wide range of demonstrations. In the next section, we demonstrate and evaluate the capabilities of our approach in a complex multi-step learning from demonstration setting.

\section{Experimental evaluation}

\begin{figure}
    \centering
    \subfigure[]{
    \label{stations.fig}
    \includegraphics[trim={0 0.09cm 0 0},clip,width=0.41\linewidth]{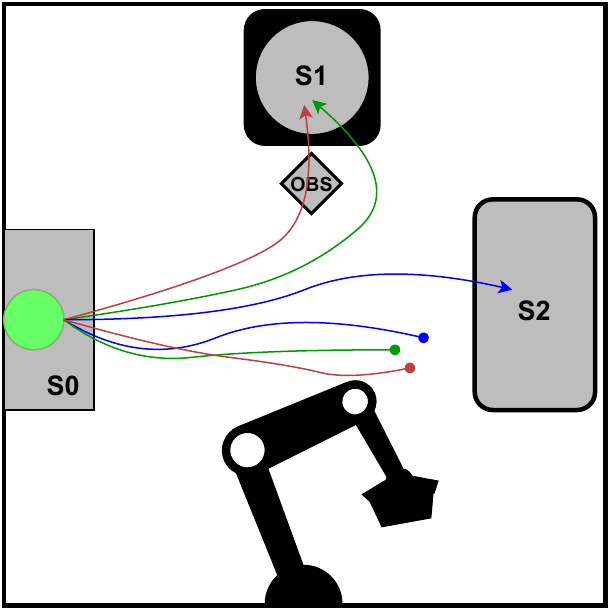}}
    \subfigure[]{
    \includegraphics[width=0.525\linewidth]{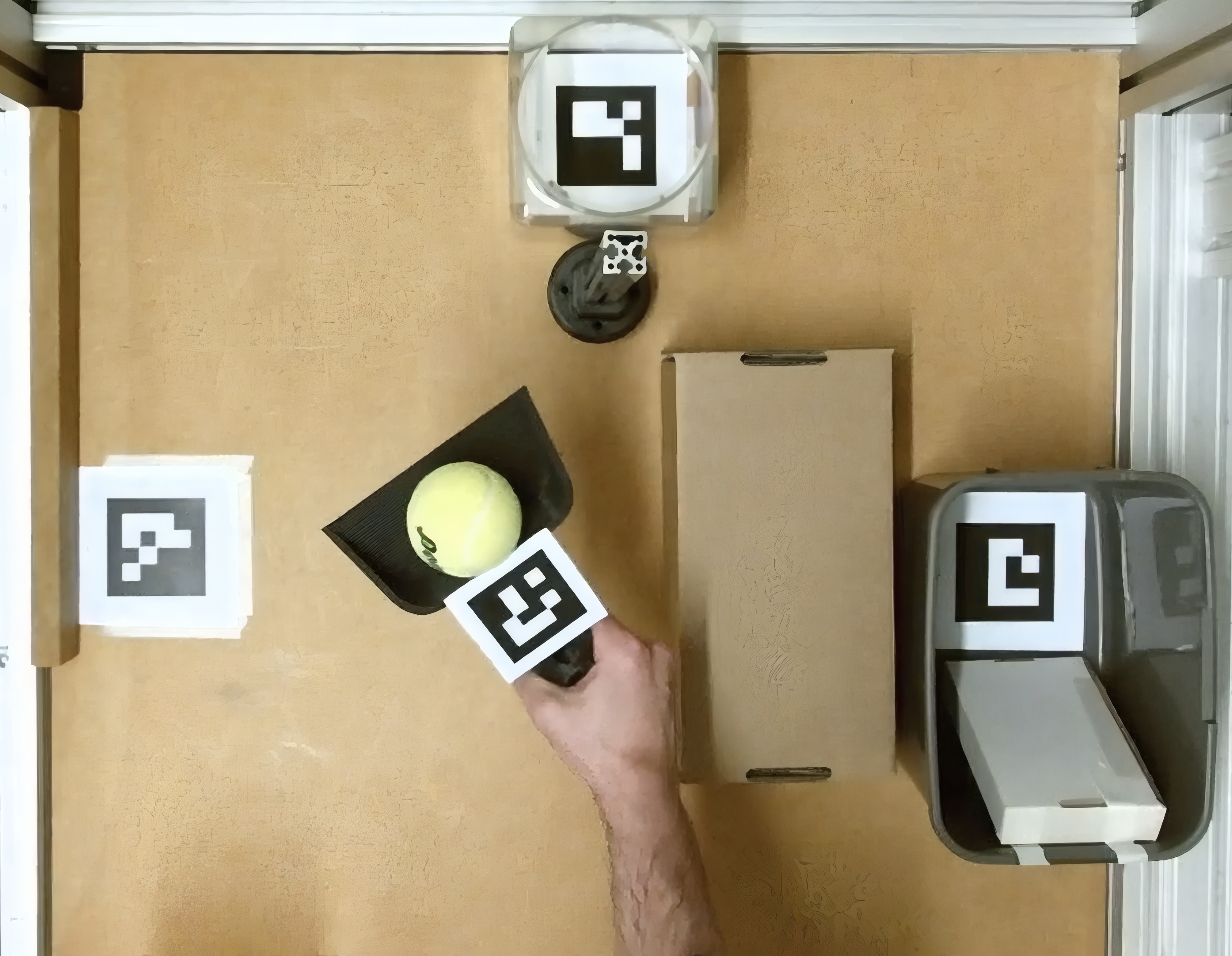}}
    \label{demo.fig}
    \caption{(a) Experimental task: The robot scoops a ball from S0, and transports it to S1 or S2. Trajectories: O1 (red), O2 (blue), and O3 (green) with an obstacle near S1.
    (b) Snapshot from a video demonstration showing O3.}
    \label{fig:stations}

\end{figure}

We implement our approach by building upon the DMP implementation from~\cite{GINESI2021103844} and the \textit{BehaviorTree.CPP}\footnote{\url{https://www.behaviortree.dev/}} library. 
For evaluation, we assess our methodology using a Franka Emika Panda 7-DOF manipulator using a Cartesian impedance controller and the experimental setup shown in Fig.~\ref{fig:setup:demo}. 
We choose a dynamic manipulation task, where the robot is supposed to slide a rigidly attached scoop under a target test object (a tennis ball), lift it up, and transport it to one of two drop areas.  
The open sides of the shovel make balancing the tennis ball a difficult task, as it easily rolls off even during small movements.

\subsection{Demonstration data collection}
To collect the demonstrations, we utilize a Microsoft Kinect v2 sensor and affix ArUco markers~\cite{aruco} to various locations, including different stations and surfaces, to track occupancy. 
Additionally, we attach a marker to the handle of the shovel to record its pose, which later represents the end-effector pose for the robot.

We collect vision-only demonstrations of a human using a shovel to perform one of three operations, all starting from the same initial position. 
Each operation consists of picking up the tennis ball with the shovel, transporting it while maintaining appropriate shovel orientation, and placing it in one of the two goal stations.
Operation 1 (O1) entails the placement of the object into station 1, while Operation 2 (O2) occurs when station 1 is already occupied, leading to the placement of the object in station 2.
Operation 3 (O3) consists of placing the ball into station 1 with the added difficulty of having an obstacle in front of the station, requiring a different motion to avoid the obstacle.
Demonstration data collected includes the tracked pose of the shovel and four engineered binary features, which represent the conditions that determine the overall scenario logic. 
We track the occupancy of start (S0), station 1 (S1), station 2 (S2), and the presence of an obstacle at station 1 (OBS), using these as state space observations.
Fig.~\ref{fig:stations} provides an overview of the experimental setup.

We follow the DMP and BT learning strategies described in Sec.~\ref{proposed_method} and perform a DMP grid search for the number of Gaussian basis functions $N_{\psi}$, (ranging from 10 to 100 with a step size of 10) and the constant of the canonical system $\alpha$ (ranging from 1 to 20 with a step size of 1). 
Due to the absence of correctness guarantees following pruning, we automatically select the BT with the lowest pruning level that closely correlates with the generated DT rules.

We tick the BT continuously until the goal is reached (\textit{Success}) or an unrecoverable failure occurs (e.g., collision). 
To enhance reactivity, ticks are triggered immediately after the root returns a status, resulting in a variable update frequency based on the number of nodes processed. 
We then evaluate our approach in increasingly complex scenarios.

\subsection{Single demonstration} \label{sec:single_demo}
To demonstrate the capability of learning from a single demonstration using our method, we create individual BTs from each of the operations previously described and assess the execution success rate of the learned policies across 25 trials. 
In all three scenarios our method learned a BT composed of three DMPs.
On average, the learning process for DMPs and BT structure took 14.4 and 12.9 seconds respectively, resulting in the robot replicating the demonstration within a minute of being shown the original motion.
The results are shown in Table \ref{tab:combined_experiment1}.

For the BT corresponding to Operation 1, the success rate was 84\%, as in 4 of the trials the ball bounced off the station boundary. 
In this case, failures can be attributed to the unsuitable orientation of the shovel, which was positioned too close to horizontal. 
Upon closer inspection, we note that the DMP controller tracking error along the orientation dimension was negligible (see Tab.~\ref{tab:combined_experiment1}). 
Consequently, we attribute this issue to pose estimation errors in the ArUco marker during the demonstration. 
While in the majority of the evaluations these errors were not crucial to the final task success, in some cases the stochasticity of the dynamics of the ball during the placement phase resulted in failures.

\begin{table}[t]
\centering
\caption{Results of Single Demonstration Experiments and DMP Controller Tracking Errors: Translation errors (meters), orientation errors (radians), and success rates over 25 trials for each operation.}
\label{tab:combined_experiment1}
\begin{tabular}{@{}llrlrr@{}}
\toprule
\multicolumn{1}{c}{\multirow{2}{*}{\textbf{Operation}}} &
  \multicolumn{2}{c}{\textbf{Translation}} &
  \multicolumn{2}{c}{\textbf{Orientation}} &
  \multicolumn{1}{c}{\multirow{2}{*}{\textbf{\begin{tabular}[c]{@{}c@{}}Success\\ rate\end{tabular}}}} \\ \cmidrule(lr){2-5}
\multicolumn{1}{l}{} & \multicolumn{1}{r}{Med} & Std   & \multicolumn{1}{r}{Med} & Std       & \multicolumn{1}{l}{}    \\ \midrule
(O1)\textit{ Place S1 }       & 0.013                   & 0.026 & 0.073                   & 0.124     & 84\%                    \\ \midrule
(O2) \textit{Place S2 }       & 0.003                   & 0.012 & 0.029                   & 0.059     & 100\%                   \\ \midrule
(O3)\textit{ Place S1 w/OBS} & 0.007                   & 0.040 & 0.037                   & 0.135     & 100\%                   \\ \midrule
\multicolumn{5}{r}{\textbf{Overall}}                                             & \textbf{94.7\%} \\
\end{tabular}%
\end{table}
For the BTs representing the other two operations, O2 and O3, the learned policies were notably more competent, successfully completing the task in all trials.  
When considering the overall performance across all experiments, the success rate averaged at 94.67\%, demonstrating a consistent level of proficiency in executing tasks across different scenarios.

\subsection{Combined demonstrations} \label{combined_experiments}

To demonstrate the adaptability of our methodology in learning from heterogeneous demonstrations, we utilize the previously learned DMPs and their associated segment demonstrations from Sec.~\ref{sec:single_demo} to construct a new BT that encapsulates all learned behaviors.
In particular, we merge all the $D$ datasets from each operation into a unified dataset and then proceed with stage 3 (BT learning) of our proposed approach. Given the DMP controllers and state variables from the previous section, the BT generation process in this scenario took 17.2 seconds.

\begin{table}[t]
\centering
\caption{Results - Combined experiments}

\label{tab:experiment2-1}

\begin{tabular}{@{}lccrr@{}}
\toprule
\textbf{Operation}&
\textbf{Disturbance} &
\textbf{New obj.} &
\textbf{\# Trials} &
\textbf{Success rate}\\
\midrule
\multirow{2}{*}{(O1) \textit{Place S1}}                           & S1 occupied    & O2 & 10 & 100\%                \\ \cmidrule(l){2-5} 
                                                         & OBS            & O3 & 10 & 100\%                \\ \midrule
\multirow{2}{*}{(O2) \textit{Place S2}}                           & S1 free        & O1 & 10 & 100\%                \\ \cmidrule(l){2-5} 
                                                         & S1 free \& OBS & O3 & 10 & 100\%                \\ \midrule
\multirow{2}{*}{(O3) \textit{Place S1 w/OBS}}                     & OBS free       & O1 & 10 & 90\%                 \\ \cmidrule(l){2-5} 
                                                         & S1 occupied    & O2 & 10 & 100\%                \\ \midrule
\multicolumn{3}{r}{\textbf{Overall}}                                           & 60 & 98.33\%             
\end{tabular}%
\end{table}

In this experiment, we artificially introduce a disturbance during the normal execution of one of the three operations to simulate real-world challenges and combine the picking of tennis ball and an unseen object during demonstrations, a rubber duck.
Disturbances are chosen such that they interrupt the current Operation and require a change of the objective and a switch to one of the other two Operations. 
For example, under Operation 1 (see Fig~\ref{fig:stations}), a possible disturbance is changing the condition that S1 is occupied, forcing the BT to change objective and continue the execution with the DMPs associated to Operation 2.

In Table~\ref{tab:experiment2-1}, we present the success rate of continuing with the operation after introducing a disturbance during the execution of the first DMP for each of the operations, corresponding to the initial scooping and lifting of the object.
Disturbances in later phases are not evaluated, as recovery is limited to halting execution.
Results show a 90–100\% success rate in adapting to disturbances, demonstrating our method's robustness and adaptability to real-world challenges and dynamic objectives.

\subsection{Editing learned behaviors}

One of the main disadvantages of LfD methods based on directly copying the behavior of the teacher is that it is hard to modify them with fallback behaviors to address failures not seen during demonstration. 
Our approach addresses this limitation by allowing a human designer to quickly and easily modify the learned behavior. 
To demonstrate the capacity of our approach to modify learned behaviors, we manually integrate a recovery behavior into the learned BT from Sec.~\ref{combined_experiments}.
This recovery behavior involves sweeping the work table with the shovel in response to a scenario where the ball has fallen mid-execution of a DMP. 
To facilitate this, we introduce a new engineered feature: the occupancy of the shovel, by tracking an ArUco marker attached to the surface of the shovel blade.

\begin{figure}[t]
    \centering
    \includegraphics[width=0.32\linewidth]{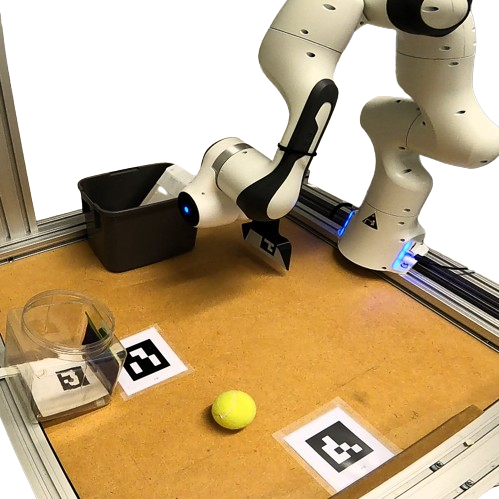}
    \includegraphics[width=0.32\linewidth]{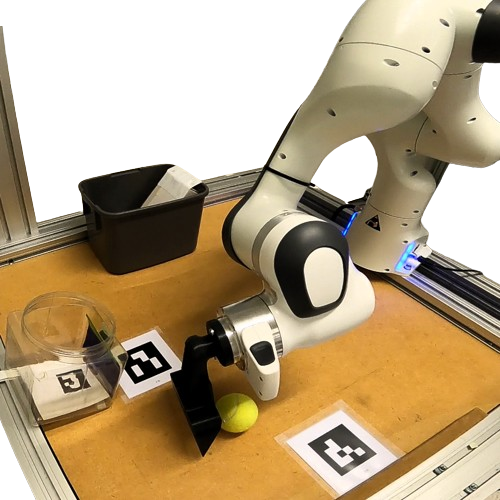}
    \includegraphics[width=0.32\linewidth]{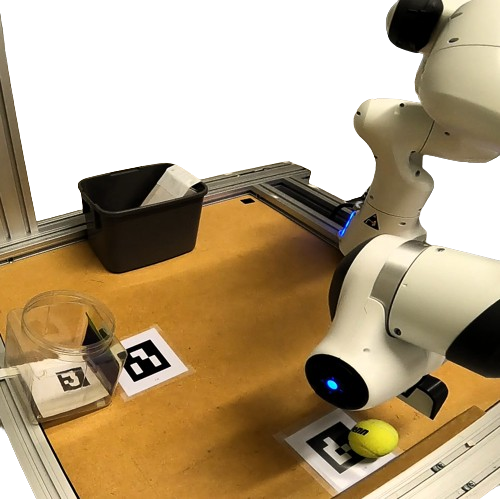}
    \caption{Recovery behavior: If the ball falls during a DMP execution, the robot transitions to sweeping the table with a shovel to return the ball to the start position.}
    \label{fig:recovery}
\end{figure}

In our experiments, we subjected the BT to react to failure scenarios across 20 trials, similar to those in Fig.~\ref{fig:recovery}. 
Our approach consistently responded to these failures in all trials, thereby demonstrating the effectiveness of our modular BT approach in modifying behaviors to manage disruptions and seamlessly transition between different control strategies.

\section{Limitations and Future Work}
One trade-off we did not explore in detail in this paper stems from the granularity of learned actions. 
While the proposed approach produces competent BTs, it does not necessarily optimize the number of action nodes used —a factor that can enhance efficiency and interpretability~\cite{gugliermo2024evaluating}.
Moreover, the parameters and structure of DMPs do not directly map to intuitive action labels, thus diminishing the overall interpretability of the BT policy.
Additionally, the choice of trajectory splitting criteria significantly influences tree structure and performance; exploring alternative criteria based on task-specific heuristics or data-driven approaches could yield better results but requires further validation. 
Lastly, scalability becomes a concern as tasks grow in complexity, requiring careful design to balance efficiency, reactivity, and system simplicity.

Future research will address the above limitations and extend the framework to incorporate online learning mechanisms and human feedback loops to facilitate policy adaption and improvement.
Finally, investigating techniques to handle uncertainty and variability in human demonstrations would contribute to the robustness and reliability of the learned robot policies.

\section{Conclusion}
In conclusion, our proposed approach presents a novel solution to the challenge of learning interpretable and adaptable robot policies from demonstration. 
By combining BTs for high-level decision-making with DMPs for low-level motion generation, we achieve a framework that can learn both the task structure and associated actions concurrently from a single demonstration, as well as combine multiple demonstrations in a coherent high-level policy.

We evaluate our approach in a real-world scenario of teaching a robot arm complex manipulation tasks and demonstrate that our method can successfully replicate and combine different demonstrations into cohesive actions.
Moreover, we show that our framework allows for manual augmentation and modification, enhancing its adaptability and usability in practical applications.

\begin{credits}
\subsubsection{\ackname}
This work was supported in part by Industrial Graduate School Collaborative AI \& Robotics (CoAIRob), in part by the Swedish Knowledge Foundation under Grant Dnr:20190128, and the Knut and Alice Wallenberg Foundation through Wallenberg AI, Autonomous Systems and Software Program (WASP).
\subsubsection{\discintname}
The authors have no competing interests to declare that are relevant to the content of this article.
\end{credits}

\bibliographystyle{splncs04}
\bibliography{bibliography}
\end{document}